%% file: colm2026_conference.tex
\documentclass{article} 
\usepackage[final]{colm2026_conference}

\usepackage{microtype}
\usepackage{hyperref}
\usepackage{url}
\usepackage{booktabs}

\usepackage{booktabs}
\usepackage{amsmath}
\usepackage{colortbl} 
\usepackage[table]{xcolor} 
\usepackage{multicol}
\usepackage{multirow}
\usepackage{import}
\usepackage[ruled,vlined]{algorithm2e}
\usepackage{wrapfig}
\definecolor{LightBlue}{rgb}{0.9, 0.95, 1.0} 
\definecolor{LightPink}{rgb}{1.0, 0.9, 0.9} 
\definecolor{LightGreen}{rgb}{0.9, 1.0, 0.9}
\usepackage{tcolorbox}
\usepackage{listings} 
\usepackage{enumitem}
\usepackage{subcaption}
\tcbuselibrary{breakable, skins}

\newtcolorbox{promptbox}[1]{
    colback=gray!5!white,
    colframe=gray!75!black,
    fonttitle=\bfseries,
    title=#1,
    breakable,
    enhanced,
    after skip=1cm
}

\usepackage{lineno}

\definecolor{darkblue}{rgb}{0, 0, 0.5}
\hypersetup{colorlinks=true, citecolor=darkblue, linkcolor=darkblue, urlcolor=darkblue}

\usepackage{fontawesome5}

\definecolor{linkcolor}{RGB}{0, 80, 150}
\hypersetup{
    colorlinks=true,
    linkcolor=linkcolor,
    filecolor=linkcolor,
    urlcolor=linkcolor,
    citecolor=linkcolor,
}
\title{CARV: A Diagnostic Benchmark for Compositional Analogical Reasoning in Multimodal LLMs}

\author{Yongkang Du, Xiaohan Zou, Minhao Cheng, Lu Lin\\
Pennsylvania State University\\
\texttt{\{ybd5136,xfz5266,mmc7149,lulin\}@psu.edu} \\
}

\begin{document}

\ifcolmsubmission
\linenumbers
\fi

\maketitle
\begin{center}
    \vspace{-0.2cm}
    \href{https://github.com/YongkDu/CARV}{\color{black}{\faGithub} \ \textbf{Code}} 
    \quad \quad \quad 
    \href{https://huggingface.co/datasets/duyongka/CARV}{\color{black}{\faDatabase} \ \textbf{Dataset}}
    \vspace{0.3cm}
\end{center}
\input{sec/0_abstract}
\section{Introduction}

\input{sec/1_intro}

\section{Related Work}

\input{sec/2_relatedwork}
\section{Compositional Analogical Reasoning}

\input{sec/3_method}
\section{Dataset Construction}
\input{sec/4_dataset}
\section{Experiment}
\input{sec/5_experiment}
\input{sec/6_analysis}
\section{Conclusion}
\input{sec/7_conclusion}


\section*{Limitation}
Our study has several limitations that point to promising directions for future work. First, the CARV dataset is built from a constrained set of synthetic scenes and a limited vocabulary of objects and furniture. This controlled design may not fully reflect the diversity of analogical reasoning. Broadening CARV to cover more visual domains would help verify whether our findings generalize beyond the current benchmark scope.
Second, we do not investigate whether targeted training could address the decomposition bottleneck identified throughout our analysis. Since models mainly fail at decomposing symbolic transformation rules from raw images, a natural next step is to study whether supervised fine-tuning or reinforcement learning on decomposition-focused tasks can improve this capability. Curriculum training that gradually increases the number of atomic transformations, or auxiliary objectives that explicitly supervise intermediate symbolic representations, may be effective directions for closing this gap. We leave such training-based interventions to future work and release the CARV dataset specifically to support this line of research.

\bibliography{colm2026_conference}
\bibliographystyle{colm2026_conference}

\appendix
\section{Appendix}
\input{sec/appendix}

\end{document}

%% file: sec/0_abstract.tex
\begin{abstract}
Analogical reasoning tests a fundamental aspect of human cognition: mapping the relation from one pair of objects to another.
Existing evaluations of this ability in multimodal large language models (MLLMs) overlook the ability to compose rules from multiple sources, a critical component of higher-order intelligence.
To close this gap, we introduce \textbf{CARV} (\textbf{C}ompositional \textbf{A}nalogical \textbf{R}easoning in \textbf{V}ision), a novel task together with a 5,500-sample dataset as the first diagnostic benchmark.
We extend the analogy from a single pair to multiple pairs, which requires MLLMs to extract symbolic rules from each pair and compose new transformations.
Evaluation on state-of-the-art MLLMs reveals a striking performance gap: even Gemini-2.5 Pro achieves only 40.4\% accuracy, far below the human-level performance of 100\%.
Diagnostic analysis shows two consistent failure modes: (1) decomposing visual changes into symbolic rules, and (2) maintaining robustness under diverse or complex settings, highlighting the limitations of current MLLMs on this task. 
\end{abstract}

%% file: sec/1_intro.tex
\label{sec:intro}

\begin{wrapfigure}[]{r}{0.5\textwidth} 
    \centering
    \vspace{-15pt} 
    \includegraphics[width=\linewidth]{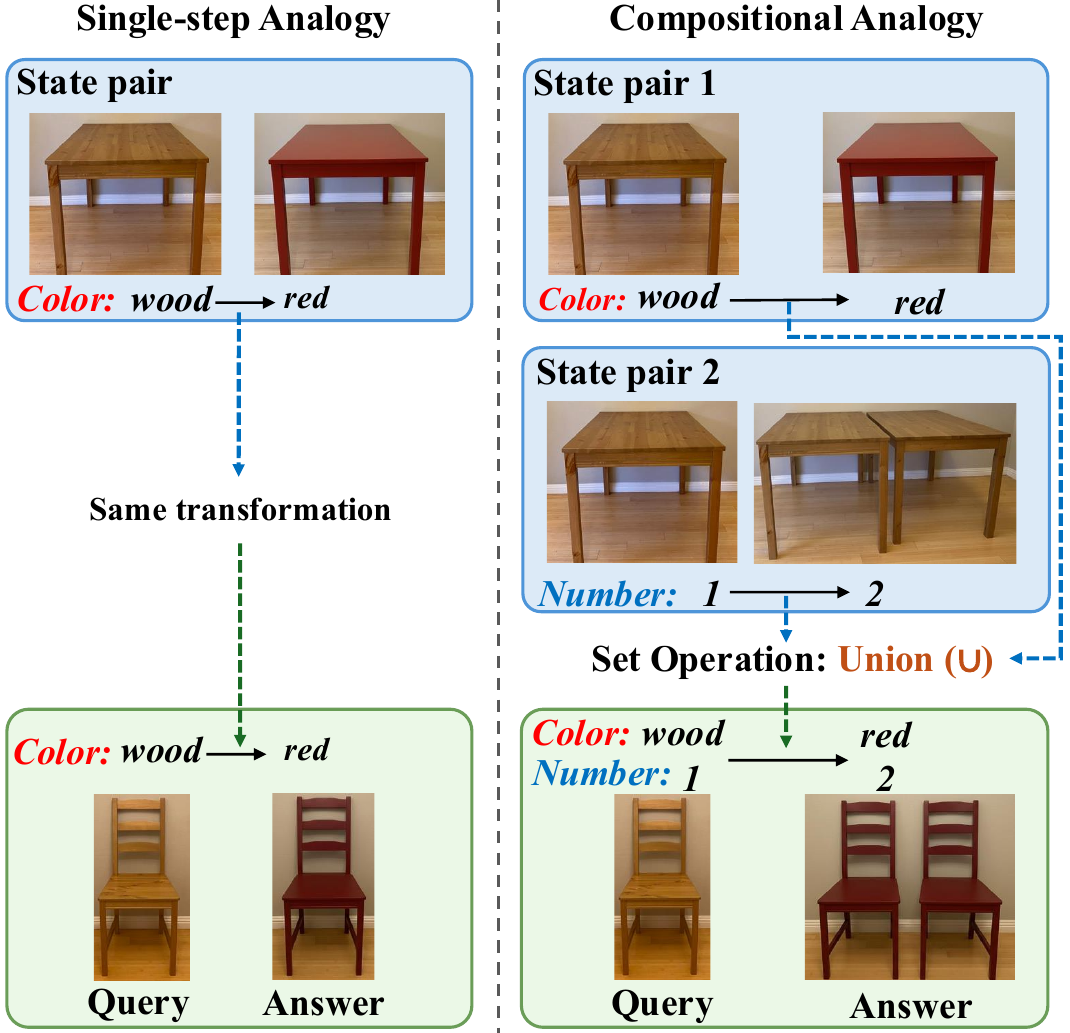}
    \caption{The single-step analogy (left) delivers one transformation (\textcolor{red}{color}) from input to target, while the compositional analogy (right) delivers the union of \textcolor{red}{color} (wood to red) and \textcolor{blue}{number} (one to two) transformations.}
    \label{fig:single_vs_comp}
    \vspace{-20pt} 
\end{wrapfigure}
The development of multimodal large language models (MLLMs)~\citep{radford2021learning, alayrac2022flamingo, comanici2025gemini} has significantly advanced visual reasoning, enabling models to tackle more complex tasks~\citep{qwen3technicalreport, chen2024internvl}.
As MLLMs approach human-level performance on many tasks, research focus has shifted toward evaluating their capacity for higher-order cognition, including complex processes such as perception~\citep{galatzer2024cognitive} and logical reasoning~\citep{ren2025large, jiang2024marvel}.

Analogical reasoning, an essential part of human cognition, enables learning from diverse contexts in an adaptive and robust way, supporting decision making and out-of-distribution generalization~\citep{mitchell2021abstraction, yiu_kiva_2025}. Current visual analogy benchmarks~\citep{yiu_kiva_2025, yilmaz_voila_2025} are mainly formulated as single-step analogy tasks (Figure~\ref{fig:single_vs_comp} left), where models only need to identify and apply the transformation rule derived from a single input pair.
However, human analogical reasoning is often sequential and constructive~\citep{sternberg1979development}. It involves combining experiential knowledge from multiple sources to solve unfamiliar problems, 
rather than focusing on a single relationship~\citep{fauconnier2008way, fauconnier2003conceptual}.
Existing visual analogy evaluation paradigms are overly simplified and therefore neglect the ability to derive novel solutions by composing and synthesizing knowledge from multiple, distinct sources. 

In this paper, we introduce \textbf{C}ompositional \textbf{A}nalogical \textbf{R}easoning in \textbf{V}ision (\textbf{CARV}).
We differ from existing visual analogy settings by evaluating a higher-order, often overlooked capability of MLLMs, namely structuring and combining knowledge from multiple sources in analogical reasoning.
As shown in Figure~\ref{fig:single_vs_comp}, existing single-step visual analogy (left) typically involves extracting and mapping a single transformation, whereas our CARV (right) goes further by demanding the ability to perform logical operations over multiple transformation sets derived from multiple image pairs.
Furthermore, CARV is developed in a controlled visual domain. As testing in open-world scenarios often contains visual noise, a controlled domain allows us to strictly isolate models' reasoning capabilities from their perception limits, which enables us to thoroughly study the failure modes.
Our contributions are summarized as follows:
\vspace{-5pt}
\begin{itemize}[leftmargin=*]
    \setlength{\itemsep}{0em}
    \item \textbf{Novel Task Formulation} We formally define the Compositional Analogical Reasoning in Vision (CARV) task. We construct a structured evaluation protocol to test a model's ability to reconstruct and apply visual transformations learned from multiple sources.
    \item \textbf{New CARV Benchmark} We introduce a comprehensive dataset spanning diverse levels of analogical reasoning complexity. CARV is carefully designed to cover orthogonal dimensions of visual transformations, including spatial and attribute transformations.
    \item \textbf{Detailed Diagnostic Analysis} We conduct extensive experiments on state-of-the-art MLLMs and find that our task is challenging for most models. To understand these limitations, we develop a fine-grained diagnosis pipeline that evaluates reasoning step by step. Our analysis reveals critical weaknesses of current MLLMs: they struggle to translate visual perceptions into symbolic, structured rules and are sensitive to increased contextual complexity.
\end{itemize}

%% file: sec/2_relatedwork.tex
\label{sec: relatedwork}
This work mainly explores whether MLLMs can perform compositional analogical reasoning. In light of this, we review two lines of research that form the basis of this work: analogical reasoning and visual composition.
\vspace{-5pt}
\paragraph{Analogical Reasoning} Analogical reasoning has long been a focus in the cognitive science~\citep{gentner1983structure, holyoak1989analogical, holyoak1997analogical}. In AI community, early studies primarily focus on lexical analogy~\citep{ushio-etal-2021-bert,fournier-etal-2020-analogies,schluter-2018-word}, i.e., king to men as queen to women, examining word pairs that share similar semantic relations. Recently, as LLMs have demonstrated advanced reasoning capabilities~\citep{liu-etal-2024-self-contradictory, du2025faircoder}, research on analogical reasoning has extended beyond lexical tasks. For example, it explores how LLMs construct analogies in real-world situations~\citep{sultan_life_2022}, applies analogical reasoning to storytelling~\citep{jiayang_storyanalogy_2023} or scientific understanding~\citep{yuan_boosting_2024}, and uses analogical reasoning as a prompting strategy~\citep{yasunaga_large_2024,qin_relevant_2025}.
In the vision domain, analogical reasoning has also received significant attention~\citep{zhang2019raven, kamath_whats_2023,zhang_multimodal_2023,guo_can_2024, lee_multimodal_2024}. The most recent benchmarks study analogical reasoning following the same format as lexical analogy, \textit{A to B as C to D}, where the input image pair consists of A and B, the query image is C, and the desired output is D. KiVA~\citep{yiu_kiva_2025}, draws inspiration from visual analogy tasks for children, employing basic image transformations (i.e., rotation and flipping) to construct the input image pair. VOILA~\citep{yilmaz_voila_2025} incorporates diverse action scenes (number, subject, and action) in the image from which the transformation is derived.
Notably, both studies find that MLLMs struggle more with applying transformations than with identifying them. However, these prior approaches are limited by task formats that rely on a single source, a constraint that our work aims to address. Instead of revisiting issues in applying transformations, we reveal the bottleneck of composition ability of current MLLMs.
\paragraph{Visual Composition} Compositional ability is fundamental for MLLMs to understand complex scenes~\citep{yi2018neural, wang2025compositional} and create images with sophisticated instructions~\citep{premsri_neuro-symbolic_2025, zhu2023topnet}.
The study of visual composition was first conducted on static scene. In CLEVR~\citep{johnson_clevr_2017} and GQA~\citep{hudson_gqa_2019}, models are trained to apply object attributes and spatial relations in question answering or caption generation.
As vision-language models become popular, the research shifts toward to the image generation process~\citep{farid_what_2025, lu2023tf}, studying the composition of scene descriptions~\citep{liu_learning_2021,avidan_compositional_2022} or task instruction~\citep{gu_composition-grounded_2025}.
AVSD~\citep{chae2025decomposing} studies decomposition ability over visual comprehension tasks.
\citet{zhu2025test} shows that widely used group-based matching metrics can systematically underestimate multimodal models' compositional reasoning ability, and propose a test-time self-improvement procedure (TTM) to recover this hidden competence. 
However, previous work mainly focuses on the composition of object attributes and neglects the composition of abstract spaces or rules, which is an essential cognitive process in human creation process~\citep{fauconnier2008way, fauconnier2001conceptual}. 
To fill this gap, our work studies the composition of transformation rules observed from image pairs.

%% file: sec/3_method.tex
\label{sec:method}
In this section, we formally define compositional analogical reasoning and describe the tasks designed to evaluate MLLMs.
\subsection{Problem Formulation}
\paragraph{Preliminary} We first introduce the single-step analogy. Figure~\ref{fig:single_vs_comp} (left) describes two image pairs sharing the same transformation, denoted as $(I:I^{\prime}) :: (I_q:I_a)$. 
We formulate an image ${I}$ as a state, defined as a set of properties $p \in \mathcal{P}$ and their corresponding values $v_p\in\mathcal{V}_p$, i.e.,  $I=\{(p,v_p) \mid p\in \mathcal{P} \}$, where $v_p\in\mathcal{V}_p$.
We define an \textbf{Atomic Transformation} as a  change in the value of a single property between two states, denoted by $(p,v_p \to v_p^{\prime})$ with $v_p \neq v_p^{\prime}$.
Multiple properties may change between two states $I$ and $I^{\prime}$, and we use $T$ to denote the set of atomic transformations between them.
\begin{equation}
    T(I,I^{\prime})=\{(p,v_p \to v_p^{\prime}) \mid (p,v_p) \in I,(p,v_p^{\prime}) \in I^{\prime}\}
\end{equation}
In single-step analogy, the transformation set between the query and answer states is the same as that of the context pair, i.e., $T(I_1,I_1^{\prime})=T(I_q,I_a)$.
\paragraph{Compositional Analogy} We extend single-step analogy to multiple state pairs $(I_1: I_1^{\prime}), \dots, (I_n: I_n^{\prime}) :: (I_q:I_a)$.
Rather than directly reapplying the same transformation, the target transformation set $T(I_q,I_a)$ here is obtained by applying a set operation $O$ over $T(I_1,I_1^{\prime}), \dots, T(I_n,I_n^{\prime})$.
For example, in Figure~\ref{fig:single_vs_comp} (right), each input state pair contains a single atomic transformation: $T(I_1,I_1^{\prime})=\{(\textit{color, wood} \to \textit{red})\}$ and $T(I_2,I_2^{\prime})=\{(\textit{number, one} \to \textit{two})\}$.
The target transformation is their union $T(I_q,I_a) = T(I_1,I_1^{\prime}) \cup T(I_2,I_2^{\prime}) = \{(\textit{color, wood} \to \textit{red}), (\textit{number, one} \to \textit{two})\}$, which is later applied to the query image.


\subsection{Task Design}
\label{sec:task_design}
In our setting, the property set is defined as $\mathcal{P} = $\{subject, subject\_number, object, object\_color, spatial\_relation\}. The subject is an everyday item and the object is a piece of furniture. A typical state can be \textit{``two bottles on a red table''}.

Consistent with the previous definitions, we evaluate MLLMs under two settings.
(1) \textbf{Single-step Analogical Reasoning:} This setting serves as a reference. The model identifies the transformation $T(I_1, I_1')$ from a single image pair and applies it to the query image $I_q$. This serves as a baseline to measure the model's ability to map a transformation from a single source to a target.
(2) \textbf{Compositional Analogical Reasoning:} Consider the case of $n=2$ image pairs, i.e., $(I_1:I_1^\prime)(I_2:I_2^\prime) :: (I_q:I_a)$. The model needs to perceive a set of atomic transformations, synthesize a new target rule through a logical operation, and apply it to a query image to produce the final outcome. We consider three logical operations, \textit{Union}, \textit{Intersection}, and \textit{Difference}, $O \in \{\cup, \cap, \setminus\}$.
To evaluate different levels of generalization, we further design two task variants.
\vspace{-5pt}
\begin{itemize}[leftmargin=*]
    \setlength{\itemsep}{0em}
    \item \textbf{Shared Source Composition:} The model is given two input state pairs in which the source images are the same as the query image, i.e., $I_q = I_1 = I_2$. This setting has relatively low contextual complexity.
    \item \textbf{Different Source Composition:} The source images in the two state pairs and the query image are all different, i.e., $I_q \neq I_1 \neq I_2$. In this more general setting, the model must identify which properties change across different source contexts and compose them to apply to a new context. This requires more advanced abilities to observe and decompose transformations without being distracted by varying visual contexts.
\end{itemize}

To constrain our task to reasoning, we ask the tested model only to generate a concrete caption of the target image instead of generating the image.

%% file: sec/4_dataset.tex
In this section, we describe the image construction and data sampling process, and then provide the statistical information of the CARV dataset.

\subsection{Data Construction}

\begin{wrapfigure}{r}{0.55\textwidth}
\vspace{-15pt}
\begin{algorithm}[H]
\caption{Data Sampling for CARV}
\label{alg:sampling}
\fontsize{9.6}{12.5}\selectfont
\SetKwInput{Require}{Require}
\SetKwInput{Ensure}{Ensure}
\SetKwInput{For}{For}

\Require{Transformation Set $T_1,T_2$, Set Operation $O$, Sample Size $N$}

Initialize dataset $\mathcal{D} \leftarrow \emptyset$\\
Synthesize transformation $T=O(T_1, T_2)$

\For{$|\mathcal{D}| < N$}{
    Sample source image $I_1$ \\
    \uIf{Shared Source}{
        {Construct 1st pair:}
        $I_1^1 \leftarrow \text{Apply}(T_1, I_1)$\\
        {Construct 2nd pair:}
        $I_1^2 \leftarrow \text{Apply}(T_2, I_1)$\\
        \tcp{Query image and label}
        $I_q \leftarrow I_1, \quad  I_a \leftarrow \text{Apply}(T, I_q)$
        
        $\mathcal{D} \leftarrow \mathcal{D} \cup\{(I_1, I_1^1), (I_1, I_1^2), (I_q, I_a))\}$
    }
    \ElseIf{Different Source}{
        Sample $I_2, I_q$, where $I_q \neq I_1 \neq I_2$
        
        $I_1' \leftarrow \text{Apply}(T_1, I_1), \quad I_2' \leftarrow \text{Apply}(T_2, I_2)$
        
        $I_a \leftarrow \text{Apply}(T, I_q)$
        
        $\mathcal{D} \leftarrow \mathcal{D} \cup\{(I_1, I_1'), (I_2, I_2'), (I_q, I_a)\}$\
    }
}
\Return{$\mathcal{D}$}\
\end{algorithm}
\vspace{-15pt}
\end{wrapfigure}
To ensure a fair comparison and avoid distribution shift, we sample image pairs for different tasks from a controlled image set, generated by editing images from~\citet{kamath_whats_2023} with Gemini-2.5 Flash Image~\citep{nanobanana}. Before editing, each image can be described by a subject (an everyday item), an object (a piece of furniture), and their spatial relation.
To adapt these images to our task, we vary the number of subjects and introduce diverse object colors, and manually remove the images (2.4\%) with quality issues like overlapping or clipping.

We then sample image pairs for both single-step and compositional tasks from this image set.
To sample images for compositional analogical reasoning, we use Algorithm~\ref{alg:sampling}.
Given the nine properties in $\mathcal{P}$, which result in a wide range of transformations, balanced sampling is necessary to prevent certain transformations from dominating the dataset.
Algorithm~\ref{alg:sampling} ensures this by sampling a fixed number of examples for each operation and transformation type.
We follow the same pipeline to sample data for the single-step task (one image pair for each query). 

\subsection{Dataset Statistic}
Our dataset consists of 5,500 unique visual analogical reasoning tasks, designed to evaluate models across varying levels of compositional complexity and logical abstraction. The statistical information is summarized in Table~\ref{tab:dataset_stats}.

\paragraph{Task Distribution}
To establish a reference for standard analogy, we include \textbf{500 single-step samples}, which require mapping the transformations from one source pair to a query.
As for \textbf{compositional samples}, we start with data in which each image pair contains two atomic transformations $|T|=2$.
(1) \textbf{Shared Source:} 1,500 samples where transformations originate from the same image $I_1$. Each set operation (\textit{Union}, \textit{Intersection}, and \textit{Difference}) has 500 samples.
(2) \textbf{Different Source:} 1,500 samples where input pairs have distinct source images. Each set operation has 500 samples.
(3) \textbf{Complexity Scaling:} To test how models perform in complex settings, we provide another 2,000 samples by extending the Shared Source setting, where we scale the number of atomic transformations within each image pair, $|T|$, from 2 to 3 and 4. For $|T|=3$, we provide 1,000 samples (500 \textit{Union}, 500 \textit{Intersection}), and the same for $|T|=4$.

\begin{wraptable}{r}{0.5\textwidth}
    \vspace{-10pt}
    \centering
    \small
    \begin{tabular}{lccc}
        \toprule
        \textbf{Task} & \textbf{\#Input}& \textbf{\#Atomic} & \textbf{Count} \\ 
        \midrule
        \textbf{Single} & 1 & 2 & 500 \\
        \midrule
        \textbf{Compos.} & & & \\
         Shared &2& 2,3,4 & 3,500 \\
         Diff.& 2 &2 & 1,500 \\
        \midrule
        \textbf{Total} & & & \textbf{5,500} \\
        \bottomrule
        \end{tabular}
        \caption{Distribution of the dataset. \#Input: number of input image pairs. \#Atomic: number of atomic transformations in each pair.
        }
    \label{tab:dataset_stats}
    \vspace{-15pt}
\end{wraptable}
\paragraph{Visual Domain}
The images are grounded in a controlled setting with a property set $\mathcal{P}$ (Section~\ref{sec:task_design}) and the corresponding value sets $\mathcal{V}_p$ for each property $p \in \mathcal{P}$. Specifically, $\mathcal{V}_{\text{subject}}$ contains 9 everyday items, and $\mathcal{V}_{\text{object}} = \{ \textit{table}, \textit{chair} \}$ includes 2 furniture types. Four spatial relations are available: $\mathcal{V}_{\text{spatial\_relation}} = \{ \textit{on}, \textit{under}, \textit{left}, \textit{right} \} $. In addition, $\mathcal{V}_{\text{subject\_number}} = \{ \textit{one}, \textit{two} \} $ and $\mathcal{V}_\text{object\_color} = \{ \textit{wood}, \textit{red}, \textit{blue} \}$.

%% file: sec/5_experiment.tex
\begin{figure}[t]
    \centering
    \includegraphics[width=\linewidth]{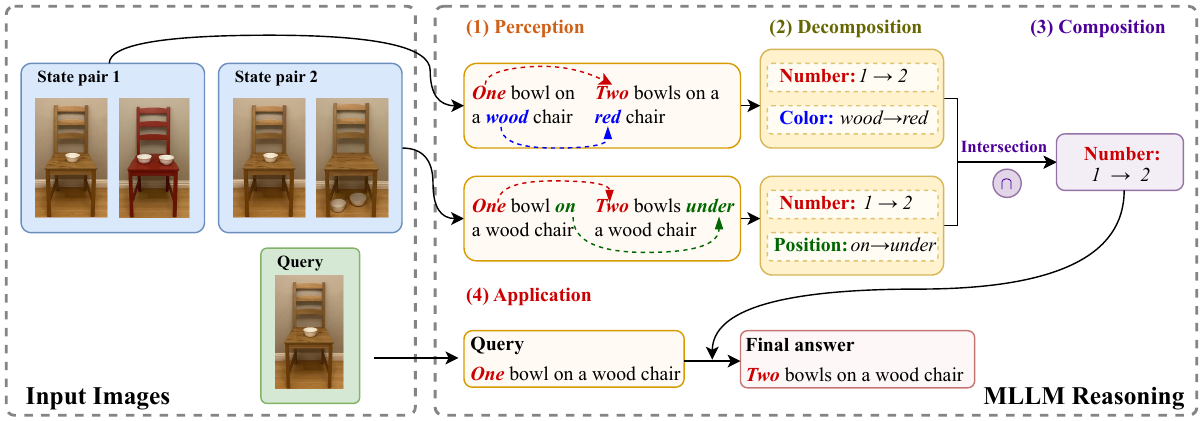}
    \caption{Overview of Diagnosis Pipeline. Given two image pairs, the model is instructed to describe the transformations, decompose the transformations into atomic transformations, perform the set operation, and apply the resulting transformation on the query image. Then we apply an evaluator model to check the correctness of each step.
    }
    \label{fig:overview}
    \vspace{-10pt}
\end{figure}
\begin{table*}[htbp] 
\vspace{-20pt}
    \centering
    { 
    \begin{tabular}{l c cc cc cc}
        \toprule
        
        \multirow{2}{*}{\textbf{Model}} 
        & \multirow{2}{*}{\textbf{Single}} 
        & \multicolumn{2}{c}{\textbf{Union}}
        & \multicolumn{2}{c}{\textbf{Intersection}}
        & \multicolumn{2}{c}{\textbf{Difference}} \\
        
        \cmidrule(lr){3-4} \cmidrule(lr){5-6} \cmidrule(lr){7-8}
        
        &
        & \multicolumn{1}{c}{\cellcolor{LightBlue}SS} & \multicolumn{1}{c}{\cellcolor{LightPink}DS}
        & \multicolumn{1}{c}{\cellcolor{LightBlue}SS} & \multicolumn{1}{c}{\cellcolor{LightPink}DS}
        & \multicolumn{1}{c}{\cellcolor{LightBlue}SS} & \multicolumn{1}{c}{\cellcolor{LightPink}DS} \\

        \midrule
        
        \textbf{Closed-source models:} & \multicolumn{7}{l}{} \\
        
        
        \cellcolor{LightGreen}{Gemini-2.5 Flash} & 75.0 & \cellcolor{LightBlue}49.4 & \cellcolor{LightPink}37.4 & \cellcolor{LightBlue}60.2 & \cellcolor{LightPink}64.2 & \cellcolor{LightBlue}41.4 & \cellcolor{LightPink}33.2 \\
        \cellcolor{LightGreen}{Gemini-2.5 Pro} & 79.2 & \cellcolor{LightBlue}51.6 & \cellcolor{LightPink}40.4 & \cellcolor{LightBlue}62.8 & \cellcolor{LightPink}67.2 & \cellcolor{LightBlue}59.4 & \cellcolor{LightPink}55.6 \\
        \cellcolor{LightGreen}{GPT-5.1} & 81.2 & \cellcolor{LightBlue}64.6 & \cellcolor{LightPink}51.0 & \cellcolor{LightBlue}75.0 & \cellcolor{LightPink}73.0 & \cellcolor{LightBlue}59.0 & \cellcolor{LightPink}62.4 \\
        GPT-4o & 44.8 & \cellcolor{LightBlue}27.2 & \cellcolor{LightPink}9.0 & \cellcolor{LightBlue}34.0 & \cellcolor{LightPink}36.0 & \cellcolor{LightBlue}25.6 & \cellcolor{LightPink}21.2 \\
        \midrule

        \textbf{Open-source models:} & \multicolumn{7}{l}{} \\
        \cellcolor{LightGreen}{Qwen3VL-8B-Thinking} & 66.7 & \cellcolor{LightBlue}33.6 & \cellcolor{LightPink}20.8 & \cellcolor{LightBlue}56.8 & \cellcolor{LightPink}59.2 & \cellcolor{LightBlue}33.3 & \cellcolor{LightPink}25.0 \\
        Qwen2.5VL-7B & 2.4 & \cellcolor{LightBlue}9.0 & \cellcolor{LightPink}2.2 & \cellcolor{LightBlue}10.0 & \cellcolor{LightPink}8.6 & \cellcolor{LightBlue}3.0 & \cellcolor{LightPink}4.2 \\
        Qwen3VL-8B & 4.8 & \cellcolor{LightBlue}5.0 & \cellcolor{LightPink}0.6 & \cellcolor{LightBlue}13.8 & \cellcolor{LightPink}8.2 & \cellcolor{LightBlue}3.2& \cellcolor{LightPink}5.4\\

        InternVL3-8B & 1.8 & \cellcolor{LightBlue}0.6 & \cellcolor{LightPink}0.2 & \cellcolor{LightBlue}5.6 & \cellcolor{LightPink}4.2 & \cellcolor{LightBlue}3.0 & \cellcolor{LightPink}3.6 \\
        \cellcolor{LightGreen}{InternVL3.5-8B} & 3.2 & \cellcolor{LightBlue}3.2 & \cellcolor{LightPink}1.6 & \cellcolor{LightBlue}7.2 & \cellcolor{LightPink}2.4 & \cellcolor{LightBlue}2.8 & \cellcolor{LightPink}0.8 \\
        Llama-3.2-11B-Vision & 1.0 & \cellcolor{LightBlue}0.6 & \cellcolor{LightPink}0.2 & \cellcolor{LightBlue}0.8 & \cellcolor{LightPink}1.4 & \cellcolor{LightBlue}0.8 & \cellcolor{LightPink}0.8 \\
        InternVL3-14B & 4.6 & \cellcolor{LightBlue}1.4 & \cellcolor{LightPink}0.0 & \cellcolor{LightBlue}7.8 & \cellcolor{LightPink}5.4 & \cellcolor{LightBlue}4.0 & \cellcolor{LightPink}2.0 \\
        Qwen2.5VL-32B & 39.0 & \cellcolor{LightBlue}18.6 & \cellcolor{LightPink}6.2 & \cellcolor{LightBlue}30.4 & \cellcolor{LightPink}32.8 & \cellcolor{LightBlue}25.4 & \cellcolor{LightPink}15.0 \\
        Qwen3VL-30B-A3B & 33.4 & \cellcolor{LightBlue}21.7 & \cellcolor{LightPink}4.2 & \cellcolor{LightBlue}18.4 & \cellcolor{LightPink}17.6 & \cellcolor{LightBlue}10.82 & \cellcolor{LightPink}9.88 \\
        
        \midrule
        Human & - & 100.0 & 100.0 & 100.0 & 100.0 & 100.0 & 100.0 \\

        \bottomrule
    \end{tabular}
    }
    \caption{Performance Comparison on Compositional Analogical Reasoning Tasks. \textit{SS} stands for same source setting and \textit{DS} stands for different source setting. The reasoning models are highlighted with green.
    }
    \label{tab:main_result}
    \vspace{-10pt}
\end{table*}
In this section, we demonstrate the experimental settings and the overall performance of current MLLMs on the proposed tasks. We first test the overall performance across different models and then apply the diagnosis pipeline to analyze the bottlenecks and failure modes.
\subsection{Experiment Setting}
\paragraph{Models} 
We evaluate state-of-the-art MLLMs, including a diverse set of closed-source and open-source models. For closed-source models, we consider GPT-5.1, GPT-4o, Gemini-2.5 Pro and Gemini-2.5 Flash. The temperature is set to $0.2$ to minimize variance in the generation process.
For open-source models, we consider QwenVL, InternVL, and Llama across various versions and parameter scales. For these models, we refer to the official configurations provided on the HuggingFace\footnote{https://huggingface.co/} site. The experiments for open-source models are conducted on NVIDIA A100 GPUs. For the human test, we recruit three participants from different backgrounds, and the instructions for the participants are the same as those used in direct prompting.

\paragraph{Prompting Methods} 
We design two prompting methods. (1) \textit{Direct Prompting}: Models are provided with image pairs and the query image together with simple but clear task instructions to predict the answer image $I_a$.
(2) \textit{Diagnosis Prompting} (Figure~\ref{fig:overview}): We divide the task into four subtasks to better analyze the reasoning process. Perception, captioning the images and describe the transformation. Decomposition, decomposing the transformation into atomic transformations. Composition, synthesizing new transformations with logical operations. Application, applying the transformations to the query image and captioning the resulting image. Models are prompted to finish the task by sequentially solving the subtasks. We apply Direct Prompting to evaluate performance on the single-step and compositional tasks and Diagnosis Prompting to understand the bottleneck. All prompts used in this work are included in Appendix~\ref{sec:prompt_temp}.

\paragraph{Evaluation} The ground-truth target images are sampled during the data sampling stage. Since models' responses may vary in linguistic phrasing, we instruct GPT-4o to evaluate the correctness of each generated caption by comparing it with the ground-truth image. To ensure reliable evaluation results, we instruct the evaluator model that the tested model may use different wording but must not conflict with the ground truth when describing each property and value. We use accuracy as the evaluation metric, representing the ratio of tasks in which the predicted image caption matches the ground truth. We manually annotate 200 data as ground truth, and GPT-4o achieves 98\% accuracy.

\subsection{Overall Experiment Result}
We evaluate the performance of diverse MLLMs across single-step analogy and compositional tasks. The comprehensive results are summarized in Table~\ref{tab:main_result}.

\textbf{Closed-source models generally outperform open-source models.}
GPT-5.1 shows superior performance across nearly all task categories, and Gemini-2.5 series achieves competitive results on the compositional tasks. 
The majority of open-source models underperform significantly on our task, except for the QwenVL series, where Qwen3VL-8B-Thinking shows outstanding performance, underscoring the advantage of reasoning models.
\textbf{Larger models generally lead to better performance.}
For example, Qwen2.5VL-32B consistently doubles or triples the performance of its 7B counterpart. However, architectural efficiency varies; Llama-3.2-11B-Vision does not exhibit a significant advantage over smaller models from other families, indicating that the underlying vision-language backbone is also critical for the reasoning tasks.
\textbf{Reasoning models outperform the non-reasoning versions.}
Comparing GPT-5.1 with GPT-4o and Gemini-2.5 Pro with Flash, models involving deeper thinking processes get better performance.
Surprisingly, Qwen3VL-8B-Thinking not only beats models at a similar parameter scale but also beats larger models like Qwen3VL-30B.
However, InternVL3.5-8B provides only limited improvements over its non-thinking predecessors, failing to match the performance of the Qwen thinking variants.

\begin{figure}[t] 
\vspace{-1cm}
    \centering
    
    \begin{subfigure}[b]{0.48\textwidth}
        \centering
        \includegraphics[width=\linewidth]{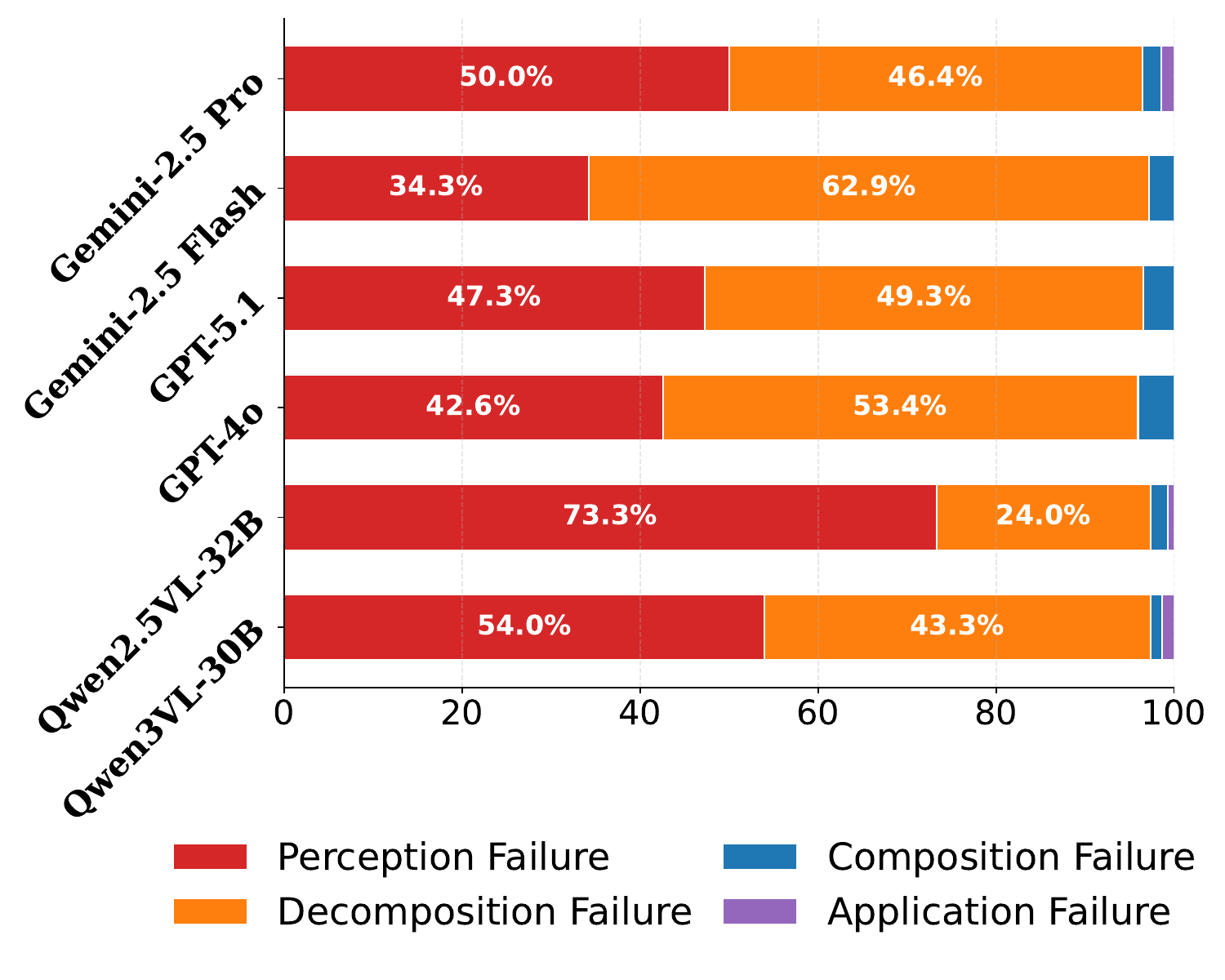}
        \caption{Error Distribution Across Models}
        \label{fig:failure_dis}
    \end{subfigure}
    \hfill 
    \begin{subfigure}[b]{0.48\textwidth}
        \centering
        \includegraphics[width=\linewidth]{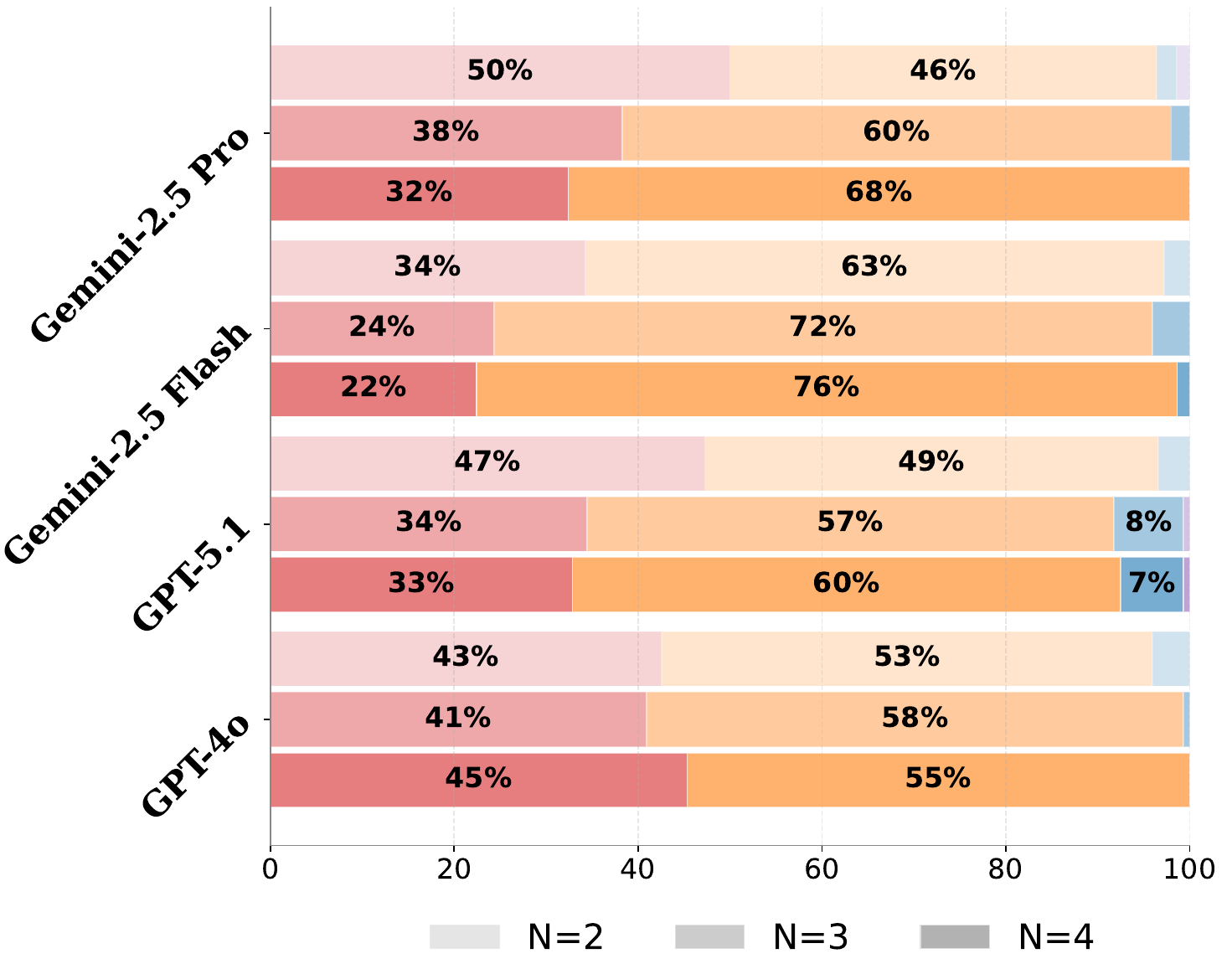}
        \caption{Error by Atomic Transformations ($N$)}
        \label{fig:error_large}
    \end{subfigure}
    
    \caption{Detailed analysis of failure distributions. \textbf{(a)} Across different models, the major bottleneck for closed-source models is decomposition, while for open-source models, it is perception. \textbf{(b)} As we scale the number of atomic transformations ($N$) in image pairs, the portion of decomposition failure significantly increases.}
    \label{fig:combined_failure_analysis}
    \vspace{-15pt}
\end{figure}

%% file: sec/6_analysis.tex
\subsection{Analysis by Subtask and Data}
\label{sec:bottleneck}
\begin{wrapfigure}{r}{0.5\textwidth}
\vspace{-1.5cm}
    \centering
    \includegraphics[width=\linewidth]{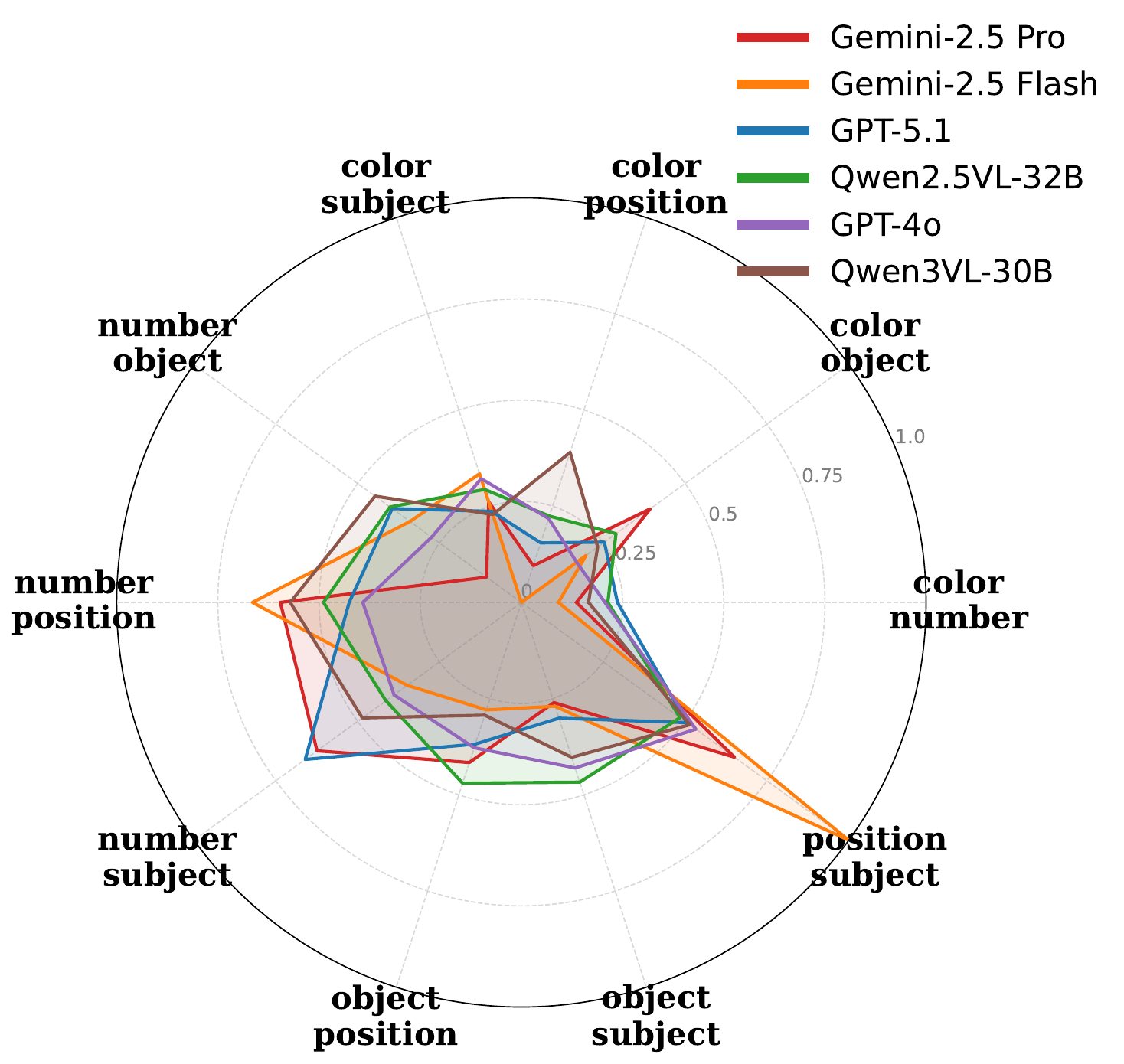}
    \caption{Failure Contribution by Property Combinations. For most models, combinations among subject, number, and position contribute most to the failure.}
    \label{fig:heatmap}
    \vspace{-20pt}
\end{wrapfigure}
In this section, we study the error distribution among subtasks and how the image contexts affect models' performance.
According to our diagnosis pipeline (Figure~\ref{fig:overview}), we instruct GPT-5.1 with the ground truth for each intermediate step and identify the step at which the tested model first fails.
To verify the reliability of the evaluator model, we perform human annotation on a subset of 120 samples. Comparing the human labels with the evaluation results, our evaluator achieves an accuracy of 0.82 and a Cohen’s Kappa~\citep{cohen1960coefficient} of 0.63, indicating high agreement.

\textcolor{blue}{\textit{\textbf{Q1: Which part of the task do the MLLMs fail at first?}}}
Results in Figure~\ref{fig:failure_dis} suggest that \textbf{the major bottlenecks are perception and decomposition}. For closed-source models, failures mostly occur at the decomposition stage, suggesting that while these models can perceive the changes, they struggle to abstract them into symbolic rules. However, for open-source models, the challenge shifts to perception.
Failures in composition and application remain relatively low, indicating that once the symbolic rules are correctly extracted, models are capable of performing the required set operations and applying them to the query image. We include failure examples in Appendix~\ref{sec:failure_cases}.

\begin{wrapfigure}{r}{0.5\textwidth}
\vspace{-10pt}
    \centering
    \includegraphics[width=\linewidth]{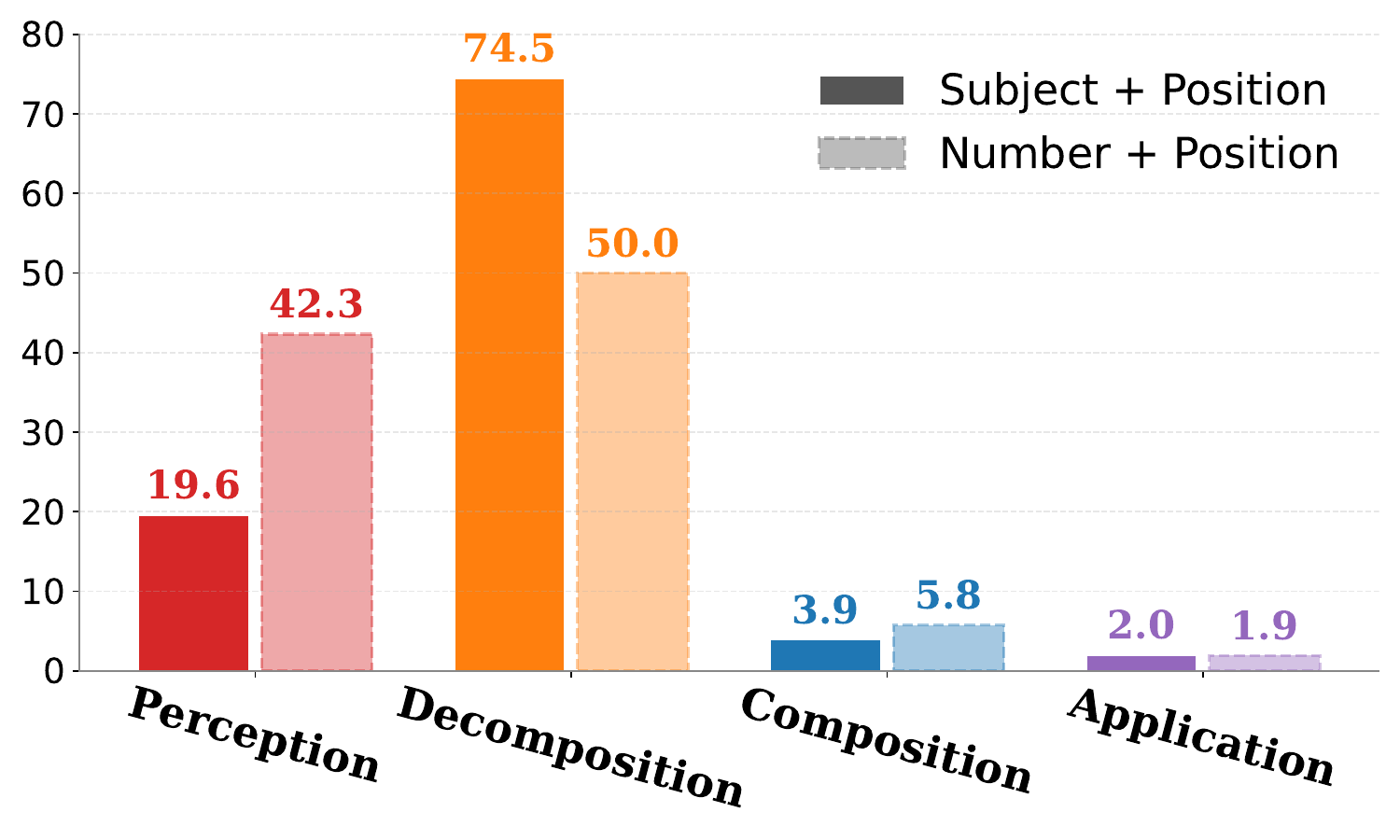}
    \caption{Error Distribution by Property Combination of Gemini-2.5 Flash. We study the most challenging combinations, the failure mainly concentrates in decomposition.}
    \label{fig:error_by_property}
    \vspace{-10pt}
\end{wrapfigure}

\textcolor{blue}{\textit{\textbf{Q2: Which property combination is most likely to lead to failure?}}} Shifting focus from the task to the image data, we investigate how specific property combinations in the transformations influence task difficulty. To achieve this goal, we employ a logistic regression model. For each task, the involved property combinations are encoded into a 10-dimensional one-hot vector (10 property combinations in total). We then use the model's correctness as the binary label.
By training the logistic regression model, we get weights that represent the contribution of each property combination toward the likelihood of failure.
Figure~\ref{fig:heatmap} shows that \textbf{the combinations among position, subject, and number contribute most significantly to failure for most models}. For instance, the combination of position and subject shows high positive weights across models, indicating that it is a primary contributor to task failure.
Furthermore, we specifically analyze the failure mode of Gemini-2.5 Flash on subject-position and number-position. Figure~\ref{fig:error_by_property} shows that the model mainly fails at decomposition for subject-position, while for number-position, it also fails at counting the subjects in the images.

\textcolor{blue}{\textit{\textbf{Q3: Are MLLMs able to generalize in different source contexts?}}}
\textbf{{Our results indicate that MLLMs generally struggle with generalization and perform worse when switching from Shared Source to the Different Source setting.}}
As shown in Table~\ref{tab:main_result}, nearly all models experience a performance drop when transitioning from Shared Source to Different Source setting.
We attribute this gap to the varying levels of abstraction required by the two settings. In the Shared Source setting, the query image $I_q$ is identical to the source image $I_1$, which allows models to potentially leverage shallow visual features. In contrast, the Different Source setting requires the model to perform a higher-order abstraction: it must decouple the transformation $T$ from the source pairs and map it to a different query image.
\begin{figure}[t]
\vspace{-20pt}
    \centering
    \includegraphics[width=\linewidth]{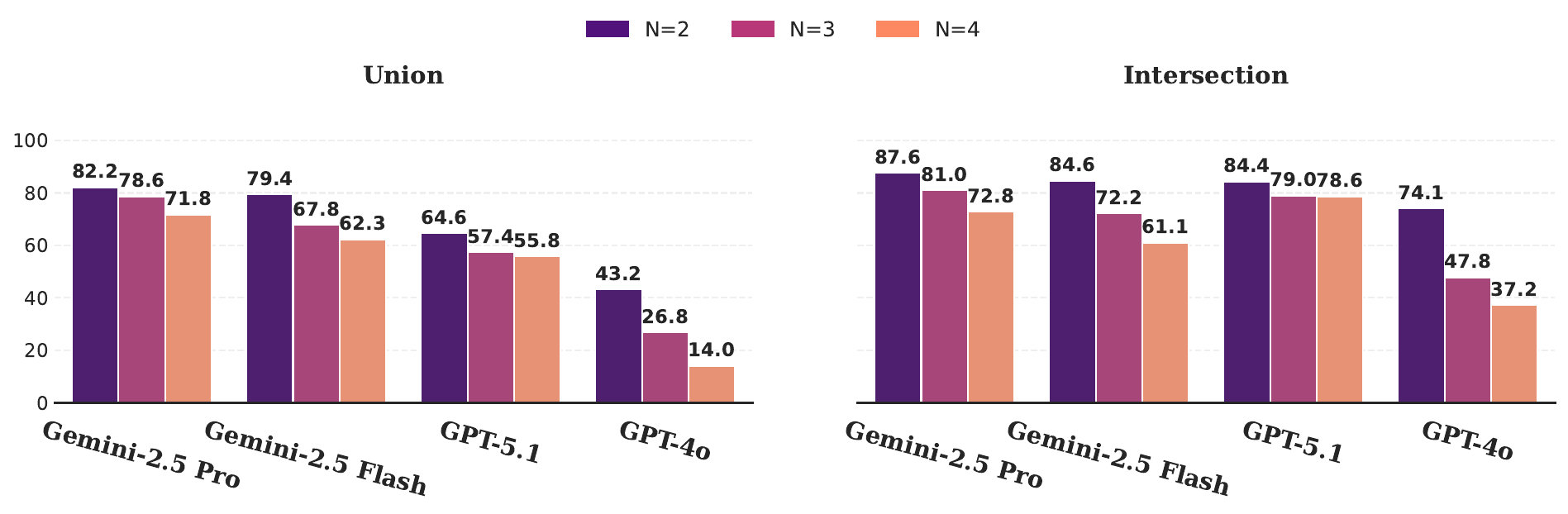}
    \caption{Accuracy by Number of Atomic Transformations. The performance drops as the number of atomic transformations ($N$) increases.}
    \label{fig:acc_atomic}
    \vspace{-15pt}
\end{figure}
\subsection{Analysis by Numerical Complexity}
In this section, we study the numerical complexity in both input transformations and output transformations after the set operation.

\textcolor{blue}{\textit{\textbf{Q4: Are MLLMs robust to scaling input transformations? If not, why?}}}
We expand the number of atomic transformations in the input pairs $|T(I,I^{\prime})|$ from $2$ to $3$ and $4$ within the Shared Source setting.
As shown in Figure~\ref{fig:acc_atomic}, \textbf{all evaluated models exhibit a consistent and significant performance decline as the number of atomic transformations increases.}
Theoretically, increasing complexity impacts both visual perception and logical composition; however, our failure analysis reveals a counter-intuitive bottleneck.
As illustrated in Figure~\ref{fig:error_large}, the proportion of {perception failure} actually decreases or remains stable as $|T|$ increases.
\textbf{However, we can observe an upward trend in decomposition failure.}
This suggests that the fundamental challenge in scaling compositional reasoning is not the visual processing, but the model's inability to disentangle and represent the increased transformations as discrete symbolic rules.

\textcolor{blue}{\textit{\textbf{Q5: Set operations lead to performance variance, is it caused by the cardinality of the output transformation set?}}}
In the previous experiment, we find that model performance varies across different set operations.
But where does the performance gap come from: the underlying logic or the output cardinality (the Union operation always increases the transformation set while the other two reduce it)?
To isolate the effect of cardinality, we design a comparison experiment involving two Union settings with identical input complexity ($|T|=2$ for each source pair) but different output cardinalities ($|T|=3$ vs. $|T|=4$).
As shown in Table~\ref{tab:cardinality_ablation}, merely increasing the cardinality does not consistently decrease performance.
\textbf{Our findings suggest that output cardinality is not the primary bottleneck; rather, task difficulty is driven by the reasoning process and the logical operations.}

\begin{wraptable}{r}{0.5\textwidth}
\vspace{-10pt}
\centering
\begin{tabular}{lcc}
\toprule
\textbf{Model} & \textbf{$|T|{= 3}$} & \textbf{$|T|{= 4}$} \\
\midrule
GPT-4o & 37.0\% & 46.0\% \\
GPT-5.1 & 85.0\% & 88.0\% \\
Gemini-2.5 Flash & 85.0\% & 74.0\% \\
Gemini-2.5 Pro & 82.0\% & 91.0\% \\
\bottomrule
\end{tabular}
\caption{Experiment on output cardinality. By increasing the output cardinality, there is no clear trend about models' performance.}
\label{tab:cardinality_ablation}
\vspace{-10pt}
\end{wraptable}

\subsection{Ablation Study for Diagnosis Prompting}
From Section~\ref{sec:bottleneck}, by applying the diagnosis prompting, we find that the main bottleneck of the compositional task lies in the decomposition stage. However, we further ask whether \textit{``the model cannot decompose the rule"} or \textit{``the model can decompose it but cannot format it"}.
We conduct an ablation study to decouple visual inference failures from formatting failures by evaluating a subset of 100 samples from the Union task across four models under three settings.
\textbf{With Format:} Models must follow the symbolic format in Diagnosis Prompting. \textbf{Without Format:} We remove all symbolic formatting constraints and allow the models to describe the visual transformations in free-form natural language. \textbf{Oracle Decomposition:} We provide the ground-truth atomic transformations in the prompt and the model is only required to perform the composition and application steps.

As detailed in Table~\ref{tab:ablation_results}, our ablation study reveals two key insights. (1) When the atomic transformations are provided, models perform nearly perfectly, indicating that they are able to compose and apply rules. {The primary failure mode is the inability to infer these rules from raw images.} (2) While removing formatting constraints yields a slight to moderate performance boost across most evaluated models, the absolute accuracy remains significantly lower than that achieved in the Oracle setting.

\begin{table}[htbp]
    \centering
    \begin{tabular}{lccc}
        \toprule
        \textbf{Model} & \textbf{With Format} & \textbf{Without Format} & \textbf{Oracle} \\
        \midrule
        GPT-4o & 42.0\% & 51.0\% & 86.0\% \\
        GPT-5.1 & 60.0\% & 78.0\% & 94.0\% \\
        Gemini-2.5 Flash & 77.0\% & 90.0\% & 99.0\% \\
        Gemini-2.5 Pro & 86.0\% & 86.0\% & 97.0\% \\
        \bottomrule
    \end{tabular}
    \caption{Task performance in ablation study. The results strengthen our claim that models struggle at the decomposition step and that the task format only slightly affects performance.}
    \label{tab:ablation_results}
    \vspace{-5pt}
\end{table}

%% file: sec/7_conclusion.tex
In this paper, we introduce Compositional Analogical Reasoning in Vision (CARV), a novel task designed to evaluate the higher-order reasoning capabilities of MLLMs, together with a 5,500-sample dataset.
Our experiments show that current models mainly underperform on this task and the major bottleneck is decomposition from visual transformations into symbolic rules.
Meanwhile, tested MLLMs struggle when changing the source image or increasing the number of atomic transformations in the context image pairs, indicating insufficient generalization and robustness.
Our findings suggest that achieving human-level analogical reasoning requires MLLMs to move beyond pattern recognition and toward higher-order abstraction. We hope the CARV dataset serves as a valuable resource for developing future multimodal systems.

%% file: sec/appendix.tex
\subsection{Prompt Template}
\label{sec:prompt_temp}
In this section, we show the prompt template we use in our experiment. The templates for Direct Prompting and evaluation are listed in the following tables, Table~\ref{tab:guided_prompt} shows the template for Guided Prompting, and Table~\ref{tab:dia_prompt} shows the prompt used for identifying the bottleneck of the task. 
\begin{tcolorbox}[colback=gray!5!white,colframe=gray!75!black,title=Direct Prompt, fonttitle=\bfseries]
\small
      You are an expert visual reasoning AI. Your task is to perform a visual reasoning task.

      You are given two image pairs and a query image. Each image pair consists of two images, showing a set of transformations from the first image to the second.
      Extract the transformation from the first image pair as T1 and the second image pair as T2, perform the set operation {operation} on the T1 and T2 to obtain the final transformation.
      {description}
      Then apply the final transformation to the query image to predict the resulting image.

      The final output should be a concise caption of the resulting image in JSON format.

      REQUIRED OUTPUT FORMAT

      \{\\
        "caption": your concise caption of the resulting image\\
      \}

\end{tcolorbox}

\begin{tcolorbox}[colback=blue!5!white,colframe=blue!75!black,title=Evaluation Prompt, fonttitle=\bfseries, breakable]
\small
      You are an evaluator for the image captioning task. Given the ground truth image and reference caption, please analyze the correctness of the model's final answer.

      If "under" is mentioned in the reference caption, it indicates that one object is positioned directly below another object in the image. Please pay special attention to this spatial relationship when evaluating the model's response.

      Here is the reference caption:
      ***
      \{ground truth caption\}
      ***

      Here is the model's response:
      ***
      \{model response\}
      ***

      Return your analysis in the following format:

            \{\\
              "correctness": true or false,\\
              "reason": "brief explanation of why the answer is correct or incorrect"\\
            \}

\end{tcolorbox}

\subsection{Failure Cases}
\label{sec:failure_cases}
In the diagnosis stage, we examine four parts: perception, decomposition, composition, and application. For perception failure, the model mainly fails to identify the color, number, or the position in the image. For application failure, the model mainly fails to apply the correct target transformation on the query image. Here, we give case studies on decomposition failure and composition failure, the two parts which are unique in our task.

\begin{tcolorbox}
[colback=yellow!5!white,colframe=yellow!75!black,title=Decomposition]
\small
\textbf{Model's Response}\\
**Caption of (I1)**: A wooden table with two rolls of toilet paper on the floor to the right.\\
**Caption of (I2)**: A wooden chair with two spatulas on the floor to the right.\\
**Transformation of (I1, I2)**: The table is replaced by a chair, and the toilet paper is replaced by spatulas. Both items remain on the floor to the right of the furniture.\\
**Caption of (I3)**: A wooden table with two rolls of toilet paper on the floor to the right.\\
**Caption of (I4)**: A wooden chair with two rolls of toilet paper on the chair's seat.\\
**Transformation of (I3, I4)**: The table is replaced by a chair, and the toilet paper moves from the floor to the seat of the chair.\\
**Transformation T1**: Furniture changes from table to chair; \textbf{items change from toilet paper to spatulas} on the floor to the right.\\
**Transformation T2**: Furniture changes from table to chair; \textbf{\textcolor{red}{toilet paper moves from the floor to the top of the chair}}.\\
...\\
\textbf{Reason}\\
In Transformation T1, the model correctly extract the atomic transformation by using the format \textit{[property] changes from A to B}, where the property is the item. However, for T2, the model fails to extract the abstract rule but uses the specific term to describe the position transformation, which is going to cause the reasoning failure. The correct T2 should be \textit{item's position change from on the floor to on the chair}.

\end{tcolorbox}

\begin{tcolorbox}
[colback=yellow!5!white,colframe=yellow!75!black,title=Composition Failure]
\small
\textbf{Model's Response}\\
...\\
2. **Decompose the Transformation of Each Image Pair**
\\**Transformation T1**: [Color] changes from wooden to red. [Quantity] changes from two knives to one knife.
\\**Transformation T2**: [Type] changes from chair to table. [Color] changes from wooden to red. [Quantity] remains the same (two knives).
\\3. **Derive a Final Target Transformation Rule**
\\**Target Transformation**: [Type] changes from chair to table. [Color] changes from wooden to red. \textbf{\textcolor{red}{[Quantity] should be at most two knives, as a change from two to one is also possible.}}\\
...\\
\textbf{Reason}\\
Here, T1 contains quantity transformation (from two to one) while T2 doesn't contain quantity transformation. As we doing \textit{UNION} operation, the model should include the quantity change in the target transformation instead of being uncertain.

\end{tcolorbox}
\subsection{Symbolic Reasoning}
\label{sec:symbolic}
We try to ask GPT-4o to solve the compositional analogical reasoning in neuro-symbolic way. Here, we refer the prompts in \citet{pan2023logic}. As shown in Table~\ref{tab:symbolic}, neuro-symbolic prompting is not as effective as expected.

\begin{table}[]
    \centering
    \begin{tabular}{c|cccc}
    \toprule
     Task    &I-S&I-D&U-S&U-D\\
    \midrule
    Accuracy& 37\% & 38\% & 20\% & 8\%\\
    \bottomrule
    \end{tabular}
    \caption{Performance of GPT-4o with neuro-symbolic prompting}
    \label{tab:symbolic}
\end{table}

\begin{table*}[t] 
\centering
\begin{tcolorbox}[colback=gray!5!white,colframe=gray!75!black,title=Diagnosis Prompt, fonttitle=\bfseries]
\small
      You are an expert visual reasoning AI. Your task is to perform a visual reasoning task.

      Each image pair consists of two images, showing a set of transformations from the first image to the second.

      Notes for describing the relation:

      Spatial relation must be one of the following: "on", "under", "left", "right". Notice, "under" means the subject is directly below the reference object, not diagonally. If there is diagonally relation, use "left" or "right".
      The reference object is a piece of furniture like table or chair.

      Task Instruction:

      1. Generate caption for images in image pairs (I1, I2) and (I3, I4) and describe the transformations in natural language.
          Use the format:
          **Caption of (I1)**: [Concise caption of image I1]
          **Caption of (I2)**: [Concise caption of image I2]
          **Transformation of (I1, I2)**: [Describe the differences of the two images, be specific and detailed]
          **Caption of (I3)**: [Concise caption of image I3]
          **Caption of (I4)**: [Concise caption of image I4]
          **Transformation of (I3, I4)**: [Describe the differences of the two images, be specific and detailed]\\
      2. Decompose the transformation of each image pair according to the property(s).
          Describe what changes, and from which to which. i.e. [texture] changes from [metal] to [glass].
          Use the format:
          **Transformation T1**: [Your description of the transformation]
          **Transformation T2**: [Your description of the transformation]\\
      3. Derive a final target transformation rule based on the set operation {operation} on T1 and T2.
          {description}
          When T1 or T2 contains items which indicate no change
          For example, T1 includes "[property] remains the same", and T2 includes "[property] changes from A to B".
          When the operation is UNION, only consider the transformation that indicates a change for that property in the target transformation.
          When the operation is INTERSECTION, consider that there is no change for that property in the target transformation.
            
          Describe what changes, and from which to which. i.e. [texture] changes from [metal] to [glass].
          Use the format:
          **Target Transformation**: [Your description of the final transformation]\\
      4. Apply this final transformation rule to the query image I5 to predict the resulting image.

      The final output should be a concise caption of the resulting image in JSON format.

      **REQUIRED OUTPUT FORMAT**

      \{
        "caption": your concise caption of the resulting image
      \}

\end{tcolorbox}
\caption{Template for Guided Prompting}
\label{tab:guided_prompt}
\end{table*}

\begin{table*}[t] 
\centering
\begin{tcolorbox}[colback=blue!5!white,colframe=blue!75!black,title=Evaluation Prompt, fonttitle=\bfseries]
\small
      You are an evaluator for a visual reasoning task. You will receive the response from a model. Your task is to evaluate the response according to the instruction below.

      Here is the model's response:
      \{response\}

      The evaluation tasks include:\\
      \textbf{1. Perception}: Check whether the model correctly perceives the visual changes in two image pairs (I1, I2) and (I3, I4).\\
        Instruction:\\
          (1) whether the response is correct according to the reference captions. The reference caption provides multiple details (color, number, position), the model's response should reflect these details accurately. Notice, if position in the reference caption is "right/left" but the response only says "under", consider it incorrect.
          For subject like scissors, "one scissors" is the same as "one pair of scissors" and "two scissors" is the same as "two pairs of scissors".\\
          Reference captions:
            Image1: \{caption1\}
            Image2: \{caption2\}
            Image3: \{caption3\}
            Image4: \{caption4\}\\
          If the captions are correct, then according to the caption, check\\
          (2) whether the model's transformation descriptions are correct;
        Consider the model correctly perceives the visual changes if the model correctly captions the images and describes the transformations for BOTH image pairs.\\
      \textbf{2. Decomposition}: Check whether the model correctly decomposes the transformation of each image pair according to the property(s).\\
        Instruction:\\
          (1) When describing one transformation, the model must follow [property] changes from [value A] to [value B].
              The property must be a general concept, like "position", "object", "subject", "color", "number", "furniture" or the combination of general concepts, i,.e., "object position", "object color" etc.
              Wrong examples could be: [position of the cup] changes from [on] to [under]; [object on table] changes from [cup] to [bowl].
              Correct examples could be: [object position] changes from [on] to [under]; [object color] changes from [red] to [blue].\\
          (2) Each transformation ([property] changes from [value A] to [value B]) must be atomic, describing only one single, indivisible property change.
              **Carefully check if the model bundles multiple properties into one transformation.**
              Wrong examples could be: [position] changes from [on chair] to [under table] (bundle position and object); [object] changes from [blue chair] to [red table] (bundle object and color).
              Notice, the model can describe transformations in one sentence or multiple sentences, as long as each transformation is atomic.
          If the response violates any of the above two rules, consider it incorrect.
          If each transformation satisfies the above two rules, check\\
          (3) Compare T1 and T2 with the reference transformations. The transformations in the reference must be included in the model's response.
          Reference transformations:
            Transformation T1: \{ground truth T1\}
            Transformation T2: \{ground truth T2\}
          If any transformation in the reference is missing or inconsistent in the response, consider it incorrect.\\
      \textbf{3. Composition}: Check whether the model correctly derives the target transformation rule based on the set operation on T1 and T2.
        Reference target transformation:
          Transformation T: \{target transformation\}\\
      \textbf{4. Application}: Check whether the model correctly applies this final transformation rule to the query image I5 to predict the resulting image.
        Query image: \{caption5\}
      \\
      **REQUIRED OUTPUT FORMAT**

      \{\\
        "failure stage": 1,2,3, or 4 (indicating which evaluation task the model first failed at; if all tasks are correct, return 0),\\
        "reasoning summary": A brief explanation of your evaluation.\\
      \}\\
      **Evaluate the response step-by-step and you MUST state all intermediate reasoning before giving the final answer.**
\end{tcolorbox}
\caption{Template for Diagnosis Prompting}
\label{tab:dia_prompt}
\end{table*}




\subsection{Variance Across Multiple Runs}
\label{sec:variance}
All main experiments use a fixed decoding temperature of 0.2 to minimize variance in model outputs. To empirically verify the stability of our results, we additionally evaluate a randomly sampled 200-question subset from the Union, Different Source setting across 4 independent runs, rather than repeating the full 5,500-sample benchmark, given the substantial API cost of doing so. As shown in Table~\ref{tab:variance}, the standard deviation across runs is low and the mean accuracy is consistent with the results reported in Table 2, confirming that our main findings are not an artifact of sampling noise from the generation process.

\begin{table}[h]
\centering
\begin{tabular}{lccccc}
\toprule
Model & Run 1 & Run 2 & Run 3 & Run 4 & Mean $\pm$ SD \\
\midrule
Gemini-2.5 Pro & 41.8 & 40.2 & 42.2 & 40.4 & 41.15 $\pm$ 0.86 \\
GPT-5.1 & 50.2 & 49.8 & 50.4 & 51.6 & 50.50 $\pm$ 0.67 \\
\bottomrule
\end{tabular}
\caption{Accuracy (\%) across 4 independent runs on a 200-question subset from the Union, Different Source setting. The low standard deviation confirms the stability of our main results.}
\label{tab:variance}
\end{table}

%% file: colm2026_conference.bib
@article{yasunaga_large_2024,
	title = {{LARGE} {LANGUAGE} {MODELS} {AS} {ANALOGICAL} {REASONERS}},
	language = {en},
	author = {Yasunaga, Michihiro and Chen, Xinyun and Li, Yujia and Pasupat, Panupong and Leskovec, Jure and Liang, Percy and Chi, Ed H and Zhou, Denny},
	year = {2024},
}

@inproceedings{qin_relevant_2025,
	address = {Vienna, Austria},
	title = {Relevant or {Random}: {Can} {LLMs} {Truly} {Perform} {Analogical} {Reasoning}?},
	isbn = {979-8-89176-256-5},
	shorttitle = {Relevant or {Random}},
	url = {https://aclanthology.org/2025.findings-acl.1230/},
	urldate = {2025-07-26},
	booktitle = {Findings of the {Association} for {Computational} {Linguistics}: {ACL} 2025},
	publisher = {Association for Computational Linguistics},
	author = {Qin, Chengwei and Xia, Wenhan and Wang, Tan and Jiao, Fangkai and Hu, Yuchen and Ding, Bosheng and Chen, Ruirui and Joty, Shafiq},
	editor = {Che, Wanxiang and Nabende, Joyce and Shutova, Ekaterina and Pilehvar, Mohammad Taher},
	month = jul,
	year = {2025},
	pages = {23993--24010},
}

@inproceedings{jiayang_storyanalogy_2023,
	address = {Singapore},
	title = {{StoryAnalogy}: {Deriving} {Story}-level {Analogies} from {Large} {Language} {Models} to {Unlock} {Analogical} {Understanding}},
	shorttitle = {{StoryAnalogy}},
	url = {https://aclanthology.org/2023.emnlp-main.706/},
	doi = {10.18653/v1/2023.emnlp-main.706},
	urldate = {2025-11-24},
	booktitle = {Proceedings of the 2023 {Conference} on {Empirical} {Methods} in {Natural} {Language} {Processing}},
	publisher = {Association for Computational Linguistics},
	author = {Jiayang, Cheng and Qiu, Lin and Chan, Tsz and Fang, Tianqing and Wang, Weiqi and Chan, Chunkit and Ru, Dongyu and Guo, Qipeng and Zhang, Hongming and Song, Yangqiu and Zhang, Yue and Zhang, Zheng},
	editor = {Bouamor, Houda and Pino, Juan and Bali, Kalika},
	month = dec,
	year = {2023},
	pages = {11518--11537},
}

@inproceedings{yuan_boosting_2024,
	address = {Miami, Florida, USA},
	title = {Boosting {Scientific} {Concepts} {Understanding}: {Can} {Analogy} from {Teacher} {Models} {Empower} {Student} {Models}?},
	shorttitle = {Boosting {Scientific} {Concepts} {Understanding}},
	url = {https://aclanthology.org/2024.emnlp-main.346/},
	doi = {10.18653/v1/2024.emnlp-main.346},
	urldate = {2025-11-24},
	booktitle = {Proceedings of the 2024 {Conference} on {Empirical} {Methods} in {Natural} {Language} {Processing}},
	publisher = {Association for Computational Linguistics},
	author = {Yuan, Siyu and Jiayang, Cheng and Qiu, Lin and Yang, Deqing},
	editor = {Al-Onaizan, Yaser and Bansal, Mohit and Chen, Yun-Nung},
	month = nov,
	year = {2024},
	pages = {6026--6036},
}

@inproceedings{sultan_life_2022,
	address = {Abu Dhabi, United Arab Emirates},
	title = {Life is a {Circus} and {We} are the {Clowns}: {Automatically} {Finding} {Analogies} between {Situations} and {Processes}},
	shorttitle = {Life is a {Circus} and {We} are the {Clowns}},
	url = {https://aclanthology.org/2022.emnlp-main.232/},
	doi = {10.18653/v1/2022.emnlp-main.232},
	urldate = {2025-11-24},
	booktitle = {Proceedings of the 2022 {Conference} on {Empirical} {Methods} in {Natural} {Language} {Processing}},
	publisher = {Association for Computational Linguistics},
	author = {Sultan, Oren and Shahaf, Dafna},
	editor = {Goldberg, Yoav and Kozareva, Zornitsa and Zhang, Yue},
	month = dec,
	year = {2022},
	pages = {3547--3562},
}

@inproceedings{ushio-etal-2021-bert,
    title = "{BERT} is to {NLP} what {A}lex{N}et is to {CV}: Can Pre-Trained Language Models Identify Analogies?",
    author = "Ushio, Asahi  and
      Espinosa Anke, Luis  and
      Schockaert, Steven  and
      Camacho-Collados, Jose",
    editor = "Zong, Chengqing  and
      Xia, Fei  and
      Li, Wenjie  and
      Navigli, Roberto",
    booktitle = "Proceedings of the 59th Annual Meeting of the Association for Computational Linguistics and the 11th International Joint Conference on Natural Language Processing (Volume 1: Long Papers)",
    month = aug,
    year = "2021",
    address = "Online",
    publisher = "Association for Computational Linguistics",
    url = "https://aclanthology.org/2021.acl-long.280/",
    doi = "10.18653/v1/2021.acl-long.280",
    pages = "3609--3624"
}

@inproceedings{fournier-etal-2020-analogies,
    title = "Analogies minus analogy test: measuring regularities in word embeddings",
    author = "Fournier, Louis  and
      Dupoux, Emmanuel  and
      Dunbar, Ewan",
    editor = "Fern{\'a}ndez, Raquel  and
      Linzen, Tal",
    booktitle = "Proceedings of the 24th Conference on Computational Natural Language Learning",
    month = nov,
    year = "2020",
    address = "Online",
    publisher = "Association for Computational Linguistics",
    url = "https://aclanthology.org/2020.conll-1.29/",
    doi = "10.18653/v1/2020.conll-1.29",
    pages = "365--375"
}

@inproceedings{schluter-2018-word,
    title = "The Word Analogy Testing Caveat",
    author = "Schluter, Natalie",
    editor = "Walker, Marilyn  and
      Ji, Heng  and
      Stent, Amanda",
    booktitle = "Proceedings of the 2018 Conference of the North {A}merican Chapter of the Association for Computational Linguistics: Human Language Technologies, Volume 2 (Short Papers)",
    month = jun,
    year = "2018",
    address = "New Orleans, Louisiana",
    publisher = "Association for Computational Linguistics",
    url = "https://aclanthology.org/N18-2039/",
    doi = "10.18653/v1/N18-2039",
    pages = "242--246"
}

@misc{yiu_kiva_2025,
	title = {{KiVA}: {Kid}-inspired {Visual} {Analogies} for {Testing} {Large} {Multimodal} {Models}},
	shorttitle = {{KiVA}},
	url = {http://arxiv.org/abs/2407.17773},
	doi = {10.48550/arXiv.2407.17773},
	urldate = {2025-04-16},
	publisher = {arXiv},
	author = {Yiu, Eunice and Qraitem, Maan and Majhi, Anisa Noor and Wong, Charlie and Bai, Yutong and Ginosar, Shiry and Gopnik, Alison and Saenko, Kate},
	month = mar,
	year = {2025},
	note = {arXiv:2407.17773 [cs]},
}

@misc{guo_can_2024,
	title = {Can {Multimodal} {Large} {Language} {Model} {Think} {Analogically}?},
	url = {http://arxiv.org/abs/2411.01307},
	doi = {10.48550/arXiv.2411.01307},
	urldate = {2025-04-16},
	publisher = {arXiv},
	author = {Guo, Diandian and Cao, Cong and Yuan, Fangfang and Wang, Dakui and Ma, Wei and Liu, Yanbing and Fu, Jianhui},
	month = nov,
	year = {2024},
	note = {arXiv:2411.01307 [cs]
version: 1},
}

@inproceedings{lee_multimodal_2024,
	address = {Bangkok, Thailand},
	title = {Multimodal {Reasoning} with {Multimodal} {Knowledge} {Graph}},
	url = {https://aclanthology.org/2024.acl-long.579/},
	doi = {10.18653/v1/2024.acl-long.579},
	urldate = {2025-04-16},
	booktitle = {Proceedings of the 62nd {Annual} {Meeting} of the {Association} for {Computational} {Linguistics} ({Volume} 1: {Long} {Papers})},
	publisher = {Association for Computational Linguistics},
	author = {Lee, Junlin and Wang, Yequan and Li, Jing and Zhang, Min},
	editor = {Ku, Lun-Wei and Martins, Andre and Srikumar, Vivek},
	month = aug,
	year = {2024},
	pages = {10767--10782},
}

@misc{zhang_multimodal_2023,
	title = {Multimodal {Analogical} {Reasoning} over {Knowledge} {Graphs}},
	url = {http://arxiv.org/abs/2210.00312},
	doi = {10.48550/arXiv.2210.00312},
	urldate = {2025-06-01},
	publisher = {arXiv},
	author = {Zhang, Ningyu and Li, Lei and Chen, Xiang and Liang, Xiaozhuan and Deng, Shumin and Chen, Huajun},
	month = mar,
	year = {2023},
	note = {arXiv:2210.00312 [cs]},
}

@misc{yilmaz_voila_2025,
	title = {{VOILA}: {Evaluation} of {MLLMs} {For} {Perceptual} {Understanding} and {Analogical} {Reasoning}},
	shorttitle = {{VOILA}},
	url = {http://arxiv.org/abs/2503.00043},
	doi = {10.48550/arXiv.2503.00043},
	urldate = {2025-06-23},
	publisher = {arXiv},
	author = {Yilmaz, Nilay and Patel, Maitreya and Luo, Yiran Lawrence and Gokhale, Tejas and Baral, Chitta and Jayasuriya, Suren and Yang, Yezhou},
	month = mar,
	year = {2025},
	note = {arXiv:2503.00043 [cs]},
}

@inproceedings{kamath_whats_2023,
	address = {Singapore},
	title = {What's “up” with vision-language models? {Investigating} their struggle with spatial reasoning},
	shorttitle = {What's “up” with vision-language models?},
	url = {https://aclanthology.org/2023.emnlp-main.568/},
	doi = {10.18653/v1/2023.emnlp-main.568},
	urldate = {2025-07-22},
	booktitle = {Proceedings of the 2023 {Conference} on {Empirical} {Methods} in {Natural} {Language} {Processing}},
	publisher = {Association for Computational Linguistics},
	author = {Kamath, Amita and Hessel, Jack and Chang, Kai-Wei},
	editor = {Bouamor, Houda and Pino, Juan and Bali, Kalika},
	month = dec,
	year = {2023},
	pages = {9161--9175},
}

@incollection{avidan_compositional_2022,
	address = {Cham},
	title = {Compositional {Visual} {Generation} with {Composable} {Diffusion} {Models}},
	volume = {13677},
	isbn = {978-3-031-19789-5 978-3-031-19790-1},
	url = {https://link.springer.com/10.1007/978-3-031-19790-1_26},
	language = {en},
	urldate = {2025-11-25},
	booktitle = {Computer {Vision} – {ECCV} 2022},
	publisher = {Springer Nature Switzerland},
	author = {Liu, Nan and Li, Shuang and Du, Yilun and Torralba, Antonio and Tenenbaum, Joshua B.},
	editor = {Avidan, Shai and Brostow, Gabriel and Cissé, Moustapha and Farinella, Giovanni Maria and Hassner, Tal},
	year = {2022},
	doi = {10.1007/978-3-031-19790-1_26},
	note = {Series Title: Lecture Notes in Computer Science},
	pages = {423--439},
}

@inproceedings{johnson_clevr_2017,
	address = {Honolulu, HI},
	title = {{CLEVR}: {A} {Diagnostic} {Dataset} for {Compositional} {Language} and {Elementary} {Visual} {Reasoning}},
	isbn = {978-1-5386-0457-1},
	shorttitle = {{CLEVR}},
	url = {https://ieeexplore.ieee.org/document/8099698/},
	doi = {10.1109/CVPR.2017.215},
	language = {en},
	urldate = {2025-11-25},
	booktitle = {2017 {IEEE} {Conference} on {Computer} {Vision} and {Pattern} {Recognition} ({CVPR})},
	publisher = {IEEE},
	author = {Johnson, Justin and Hariharan, Bharath and Van Der Maaten, Laurens and Fei-Fei, Li and Zitnick, C. Lawrence and Girshick, Ross},
	month = jul,
	year = {2017},
	pages = {1988--1997},
}

@inproceedings{premsri_neuro-symbolic_2025,
	address = {Albuquerque, New Mexico},
	title = {Neuro-symbolic {Training} for {Reasoning} over {Spatial} {Language}},
	isbn = {979-8-89176-195-7},
	url = {https://aclanthology.org/2025.findings-naacl.128/},
	doi = {10.18653/v1/2025.findings-naacl.128},
	urldate = {2025-11-25},
	booktitle = {Findings of the {Association} for {Computational} {Linguistics}: {NAACL} 2025},
	publisher = {Association for Computational Linguistics},
	author = {Premsri, Tanawan and Kordjamshidi, Parisa},
	editor = {Chiruzzo, Luis and Ritter, Alan and Wang, Lu},
	month = apr,
	year = {2025},
	pages = {2395--2414},
}

@inproceedings{hudson_gqa_2019,
	address = {Long Beach, CA, USA},
	title = {{GQA}: {A} {New} {Dataset} for {Real}-{World} {Visual} {Reasoning} and {Compositional} {Question} {Answering}},
	copyright = {https://doi.org/10.15223/policy-029},
	isbn = {978-1-7281-3293-8},
	shorttitle = {{GQA}},
	url = {https://ieeexplore.ieee.org/document/8953451/},
	doi = {10.1109/CVPR.2019.00686},
	language = {en},
	urldate = {2025-11-25},
	booktitle = {2019 {IEEE}/{CVF} {Conference} on {Computer} {Vision} and {Pattern} {Recognition} ({CVPR})},
	publisher = {IEEE},
	author = {Hudson, Drew A. and Manning, Christopher D.},
	month = jun,
	year = {2019},
	pages = {6693--6702},
}

@misc{liu_learning_2021,
	title = {Learning to {Compose} {Visual} {Relations}},
	url = {http://arxiv.org/abs/2111.09297},
	doi = {10.48550/arXiv.2111.09297},
	urldate = {2025-11-25},
	publisher = {arXiv},
	author = {Liu, Nan and Li, Shuang and Du, Yilun and Tenenbaum, Joshua B. and Torralba, Antonio},
	month = nov,
	year = {2021},
	note = {arXiv:2111.09297 [cs]},
}

@misc{farid_what_2025,
	title = {What {Drives} {Compositional} {Generalization} in {Visual} {Generative} {Models}?},
	url = {http://arxiv.org/abs/2510.03075},
	doi = {10.48550/arXiv.2510.03075},
	urldate = {2025-11-25},
	publisher = {arXiv},
	author = {Farid, Karim and Sahay, Rajat and Alnaggar, Yumna Ali and Schrodi, Simon and Fischer, Volker and Schmid, Cordelia and Brox, Thomas},
	month = oct,
	year = {2025},
	note = {arXiv:2510.03075 [cs]},
}

@misc{gu_composition-grounded_2025,
	title = {Composition-{Grounded} {Instruction} {Synthesis} for {Visual} {Reasoning}},
	url = {http://arxiv.org/abs/2510.15040},
	doi = {10.48550/arXiv.2510.15040},
	urldate = {2025-11-25},
	publisher = {arXiv},
	author = {Gu, Xinyi and Mao, Jiayuan and Hong, Zhang-Wei and Yu, Zhuoran and Li, Pengyuan and Joshi, Dhiraj and Feris, Rogerio and He, Zexue},
	month = oct,
	year = {2025},
	note = {arXiv:2510.15040 [cs]},
}

@article{fauconnier2001conceptual,
  title={Conceptual blending and analogy},
  author={Fauconnier, Gilles},
  journal={The analogical mind: Perspectives from cognitive science},
  volume={255},
  pages={286},
  year={2001},
  publisher={MIT Press Cambridge, MA}
}

@article{yi2018neural,
  title={Neural-symbolic vqa: Disentangling reasoning from vision and language understanding},
  author={Yi, Kexin and Wu, Jiajun and Gan, Chuang and Torralba, Antonio and Kohli, Pushmeet and Tenenbaum, Josh},
  journal={Advances in neural information processing systems},
  volume={31},
  year={2018}
}

@article{wang2025compositional,
  title={Compositional Scene Understanding through Inverse Generative Modeling},
  author={Wang, Yanbo and Dauwels, Justin and Du, Yilun},
  journal={arXiv preprint arXiv:2505.21780},
  year={2025}
}

@inproceedings{lu2023tf,
  title={Tf-icon: Diffusion-based training-free cross-domain image composition},
  author={Lu, Shilin and Liu, Yanzhu and Kong, Adams Wai-Kin},
  booktitle={Proceedings of the IEEE/CVF International Conference on Computer Vision},
  pages={2294--2305},
  year={2023}
}

@inproceedings{zhu2023topnet,
  title={Topnet: Transformer-based object placement network for image compositing},
  author={Zhu, Sijie and Lin, Zhe and Cohen, Scott and Kuen, Jason and Zhang, Zhifei and Chen, Chen},
  booktitle={Proceedings of the IEEE/CVF Conference on Computer Vision and Pattern Recognition},
  pages={1838--1847},
  year={2023}
}

@article{comanici2025gemini,
  title={Gemini 2.5: Pushing the frontier with advanced reasoning, multimodality, long context, and next generation agentic capabilities},
  author={Comanici, Gheorghe and Bieber, Eric and Schaekermann, Mike and Pasupat, Ice and Sachdeva, Noveen and Dhillon, Inderjit and Blistein, Marcel and Ram, Ori and Zhang, Dan and Rosen, Evan and others},
  journal={arXiv preprint arXiv:2507.06261},
  year={2025}
}

@misc{qwen3technicalreport,
      title={Qwen3 Technical Report}, 
      author={Qwen Team},
      year={2025},
      eprint={2505.09388},
      archivePrefix={arXiv},
      primaryClass={cs.CL},
      url={https://arxiv.org/abs/2505.09388}, 
}

@inproceedings{chen2024internvl,
  title={Internvl: Scaling up vision foundation models and aligning for generic visual-linguistic tasks},
  author={Chen, Zhe and Wu, Jiannan and Wang, Wenhai and Su, Weijie and Chen, Guo and Xing, Sen and Zhong, Muyan and Zhang, Qinglong and Zhu, Xizhou and Lu, Lewei and others},
  booktitle={Proceedings of the IEEE/CVF conference on computer vision and pattern recognition},
  pages={24185--24198},
  year={2024}
}

@article{galatzer2024cognitive,
  title={The cognitive capabilities of generative AI: A comparative analysis with human benchmarks},
  author={Galatzer-Levy, Isaac R and Munday, David and McGiffin, Jed and Liu, Xin and Karmon, Danny and Labzovsky, Ilia and Moroshko, Rivka and Zait, Amir and McDuff, Daniel},
  journal={arXiv preprint arXiv:2410.07391},
  year={2024}
}

@inproceedings{ren2025large,
  title={Do Large Language Models Mirror Cognitive Language Processing?},
  author={Ren, Yuqi and Jin, Renren and Zhang, Tongxuan and Xiong, Deyi},
  booktitle={Proceedings of the 31st International Conference on Computational Linguistics},
  pages={2988--3001},
  year={2025}
}

@book{fauconnier2008way,
  title={The way we think: Conceptual blending and the mind's hidden complexities},
  author={Fauconnier, Gilles and Turner, Mark},
  year={2008},
  publisher={Basic books}
}

@article{fauconnier2003conceptual,
  title={Conceptual blending, form and meaning},
  author={Fauconnier, Gilles and Turner, Mark},
  journal={Recherches en communication},
  volume={19},
  pages={57--86},
  year={2003}
}

@article{sternberg1979development,
  title={The development of analogical reasoning processes},
  author={Sternberg, Robert J and Rifkin, Bathsheva},
  journal={Journal of experimental child psychology},
  volume={27},
  number={2},
  pages={195--232},
  year={1979},
  publisher={Elsevier}
}

@misc{nanobanana,
  author = {Fortin, Alisa and Vernade, Guillaume and Kampf, Kat and Reshi, Ammaar},
  title = {Introducing Gemini 2.5 Flash Image, our state-of-the-art image model},
  year = {2025},
  url = {https://developers.googleblog.com/introducing-gemini-2-5-flash-image/},
  note = {Accessed: 2025-12-21}
}

@inproceedings{liu-etal-2024-self-contradictory,
    title = "Self-contradictory reasoning evaluation and detection",
    author = "Liu, Ziyi  and
      Sanyal, Soumya  and
      Lee, Isabelle  and
      Du, Yongkang  and
      Gupta, Rahul  and
      Liu, Yang  and
      Zhao, Jieyu",
    editor = "Al-Onaizan, Yaser  and
      Bansal, Mohit  and
      Chen, Yun-Nung",
    booktitle = "Findings of the Association for Computational Linguistics: EMNLP 2024",
    month = nov,
    year = "2024",
    address = "Miami, Florida, USA",
    publisher = "Association for Computational Linguistics",
    url = "https://aclanthology.org/2024.findings-emnlp.213/",
    doi = "10.18653/v1/2024.findings-emnlp.213",
    pages = "3725--3742"
}

@article{du2025faircoder,
  title={Faircoder: Evaluating social bias of llms in code generation},
  author={Du, Yongkang and Huang, Jen-tse and Zhao, Jieyu and Lin, Lu},
  journal={arXiv preprint arXiv:2501.05396},
  year={2025}
}

@article{cohen1960coefficient,
  title={A coefficient of agreement for nominal scales},
  author={Cohen, Jacob},
  journal={Educational and psychological measurement},
  volume={20},
  number={1},
  pages={37--46},
  year={1960},
  publisher={Sage Publications Sage CA: Thousand Oaks, CA}
}

@article{chae2025decomposing,
  title={Decomposing Complex Visual Comprehension into Atomic Visual Skills for Vision Language Models},
  author={Chae, Hyunsik and Yoon, Seungwoo and Park, Jaden and Chun, Chloe Yewon and Cho, Yongin and Cai, Mu and Lee, Yong Jae and Ryu, Ernest K},
  journal={arXiv preprint arXiv:2505.20021},
  year={2025}
}

@article{mitchell2021abstraction,
  title={Abstraction and analogy-making in artificial intelligence},
  author={Mitchell, Melanie},
  journal={Annals of the New York Academy of Sciences},
  volume={1505},
  number={1},
  pages={79--101},
  year={2021},
  publisher={Wiley Online Library}
}

@inproceedings{pan2023logic,
  title={Logic-lm: Empowering large language models with symbolic solvers for faithful logical reasoning},
  author={Pan, Liangming and Albalak, Alon and Wang, Xinyi and Wang, William},
  booktitle={Findings of the Association for Computational Linguistics: EMNLP 2023},
  pages={3806--3824},
  year={2023}
}

@inproceedings{zhang2019raven,
  title={Raven: A dataset for relational and analogical visual reasoning},
  author={Zhang, Chi and Gao, Feng and Jia, Baoxiong and Zhu, Yixin and Zhu, Song-Chun},
  booktitle={Proceedings of the IEEE/CVF conference on computer vision and pattern recognition},
  pages={5317--5327},
  year={2019}
}

@article{jiang2024marvel,
  title={Marvel: Multidimensional abstraction and reasoning through visual evaluation and learning},
  author={Jiang, Yifan and Zhang, Jiarui and Sun, Kexuan and Sourati, Zhivar and Ahrabian, Kian and Ma, Kaixin and Ilievski, Filip and Pujara, Jay},
  journal={Advances in Neural Information Processing Systems},
  volume={37},
  pages={46567--46592},
  year={2024}
}

@article{gentner1983structure,
  title={Structure-mapping: A theoretical framework for analogy},
  author={Gentner, Dedre},
  journal={Cognitive science},
  volume={7},
  number={2},
  pages={155--170},
  year={1983},
  publisher={Elsevier}
}

@article{holyoak1989analogical,
  title={Analogical mapping by constraint satisfaction},
  author={Holyoak, Keith J and Thagard, Paul},
  journal={Cognitive science},
  volume={13},
  number={3},
  pages={295--355},
  year={1989},
  publisher={Wiley Online Library}
}

@article{holyoak1997analogical,
  title={The analogical mind.},
  author={Holyoak, Keith J and Thagard, Paul},
  journal={American psychologist},
  volume={52},
  number={1},
  pages={35},
  year={1997},
  publisher={American Psychological Association}
}

@inproceedings{radford2021learning,
  title={Learning transferable visual models from natural language supervision},
  author={Radford, Alec and Kim, Jong Wook and Hallacy, Chris and Ramesh, Aditya and Goh, Gabriel and Agarwal, Sandhini and Sastry, Girish and Askell, Amanda and Mishkin, Pamela and Clark, Jack and others},
  booktitle={International conference on machine learning},
  pages={8748--8763},
  year={2021},
  organization={PmLR}
}

@article{alayrac2022flamingo,
  title={Flamingo: a visual language model for few-shot learning},
  author={Alayrac, Jean-Baptiste and Donahue, Jeff and Luc, Pauline and Miech, Antoine and Barr, Iain and Hasson, Yana and Lenc, Karel and Mensch, Arthur and Millican, Katherine and Reynolds, Malcolm and others},
  journal={Advances in neural information processing systems},
  volume={35},
  pages={23716--23736},
  year={2022}
}

@article{zhu2025test,
  title={Test-Time Matching: Unlocking Compositional Reasoning in Multimodal Models},
  author={Zhu, Yinglun and Zhang, Jiancheng and Tang, Fuzhi},
  journal={arXiv preprint arXiv:2510.07632},
  year={2025}
}
